\documentclass[10pt,twocolumn,letterpaper]{article}

\usepackage{cvpr}      % To produce the REVIEW version
\usepackage{graphicx}
\usepackage{amsmath}
\usepackage{amssymb}
\usepackage{algorithm2e}
\usepackage{booktabs}

\RestyleAlgo{ruled}

\usepackage[pagebackref,breaklinks,colorlinks]{hyperref}

\usepackage[capitalize]{cleveref}
\crefname{section}{Sec.}{Secs.}
\Crefname{section}{Section}{Sections}
\Crefname{table}{Table}{Tables}
\crefname{table}{Tab.}{Tabs.}

\def\cvprPaperID{7773} % *** Enter the CVPR Paper ID here
\def\confName{CVPR}
\def\confYear{2023}

\begin{document}

\title{Learning from Abstract Images: on the Importance of Occlusion in a Minimalist Encoding of Human Poses}

\author{Saad Manzur\\
University of California, Irvine\\
{\tt\small smanzur@uci.edu}
\and
Wayne Hayes\\
University of California, Irvine\\
{\tt\small whayes@uci.edu}
}

\maketitle

%%%%%%%%% ABSTRACT
\begin{abstract}
   Existing 2D-to-3D pose lifting networks suffer from poor performance in cross-dataset benchmarks. Although the use of 2D keypoints joined by ``stick-figure’’ limbs has shown promise as an intermediate step, stick-figures do not account for occlusion information that is often inherent in an image. In this paper, we propose a novel representation using opaque 3D limbs that preserves occlusion information while implicitly encoding joint locations. Crucially, when training on data with accurate three-dimensional keypoints and without part-maps, this representation allows training on abstract synthetic images, with occlusion, from as many synthetic viewpoints as desired. The result is a pose defined by limb angles rather than joint positions—because poses are, in the real world, independent of cameras—allowing us to predict poses that are completely independent of camera viewpoint. The result provides not only an improvement in same-dataset benchmarks, but a ``quantum leap" in cross-dataset benchmarks.
\end{abstract}
\vspace{-2mm} \section{Introduction}

Recent work on 3D human pose estimation from still images can be classified into two main groups: direct-from-image methods \cite{i3d_li20143d, i3d_li2015maximum, i3d_pavlakos2017coarse, i3d_pavlakos2018ordinal, i3d_sun2017compositional, i3d_tekin2016structured, i3d_zhou2016deep}, and methods that use intermediate 2D/2.5D keypoints \cite{2d3d_chen20173d, 2d3d_chen2019unsupervised, 2d3d_gcn_ci2019optimizing, 2d3d_gcn_liu2020comprehensive, 2d3d_gcn_zeng2020srnet, 2d3d_gcn_zhao2019semantic, 2d3d_gcn_zou2021modulated, 2d3d_gong2021poseaug, 2d3d_hyp_li2019generating, 25d3d_habibie2019wild}. The ease and success of 2D keypointing has resulted in a number of ``lifting’’ methods designed to transform the intermediate 2D keypoints to the 3D pose.

\begin{figure}[t]
    \centering
    \begin{subfigure}{0.3\linewidth}
        \centering
        \includegraphics[width=0.48\columnwidth]{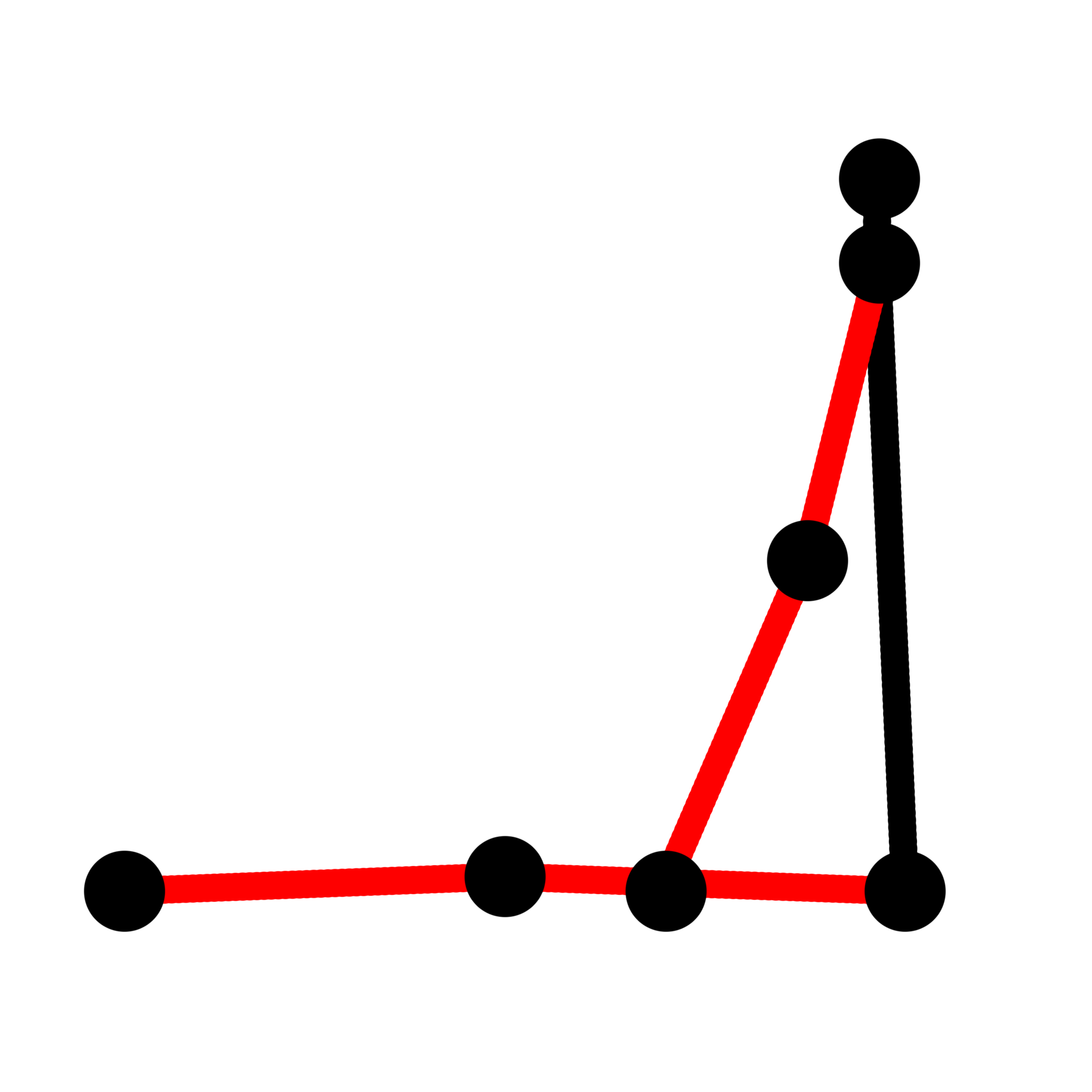}
        \caption{Left view (sitting)}
        \label{fig:left-p}
    \end{subfigure}
    \begin{subfigure}{0.3\linewidth}
        \centering
        \includegraphics[width=0.48\columnwidth]{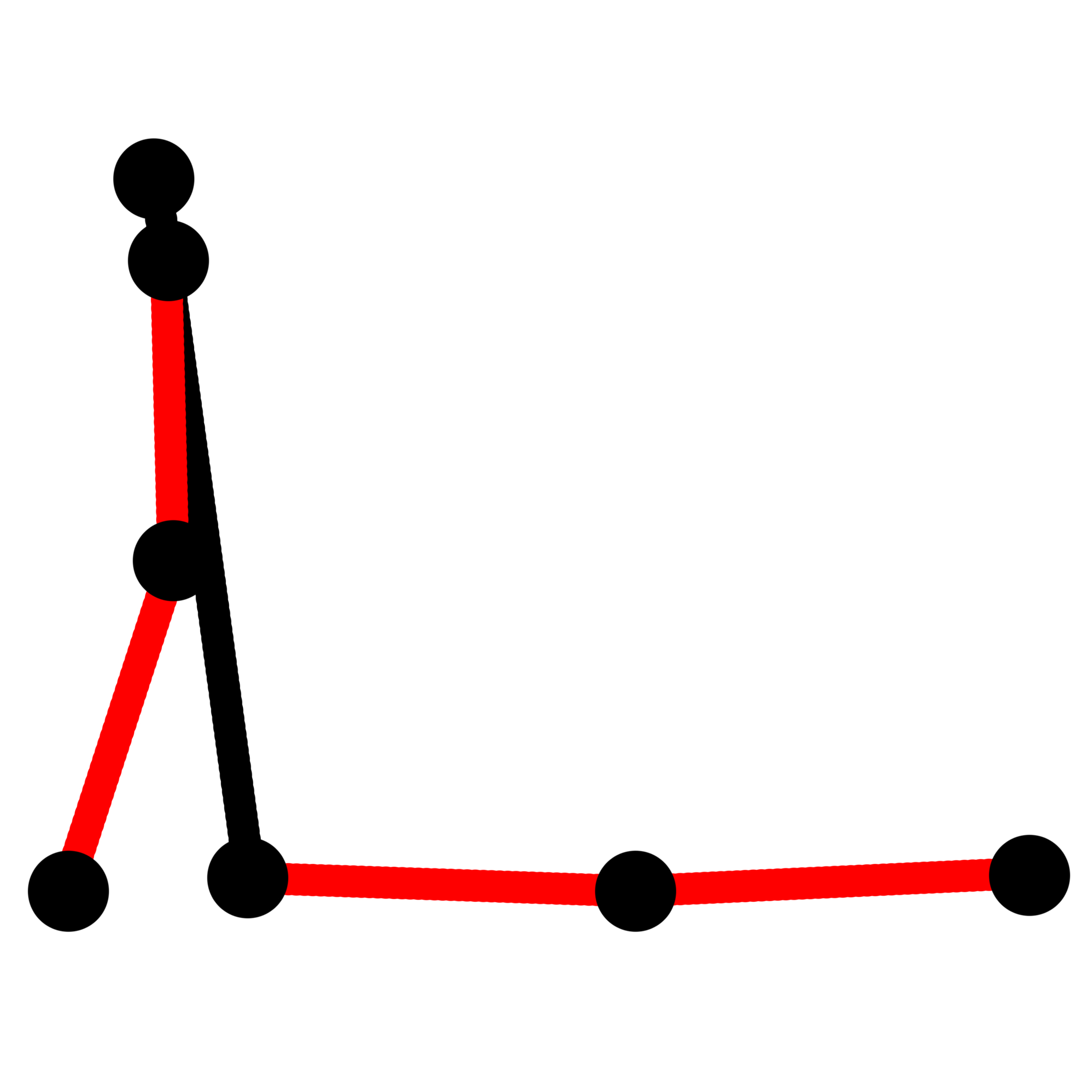}
        \caption{Left view (yoga)}
        \label{fig:left-y}
    \end{subfigure}
    \begin{subfigure}{0.3\linewidth}
        \centering
        \includegraphics[width=0.48\columnwidth]{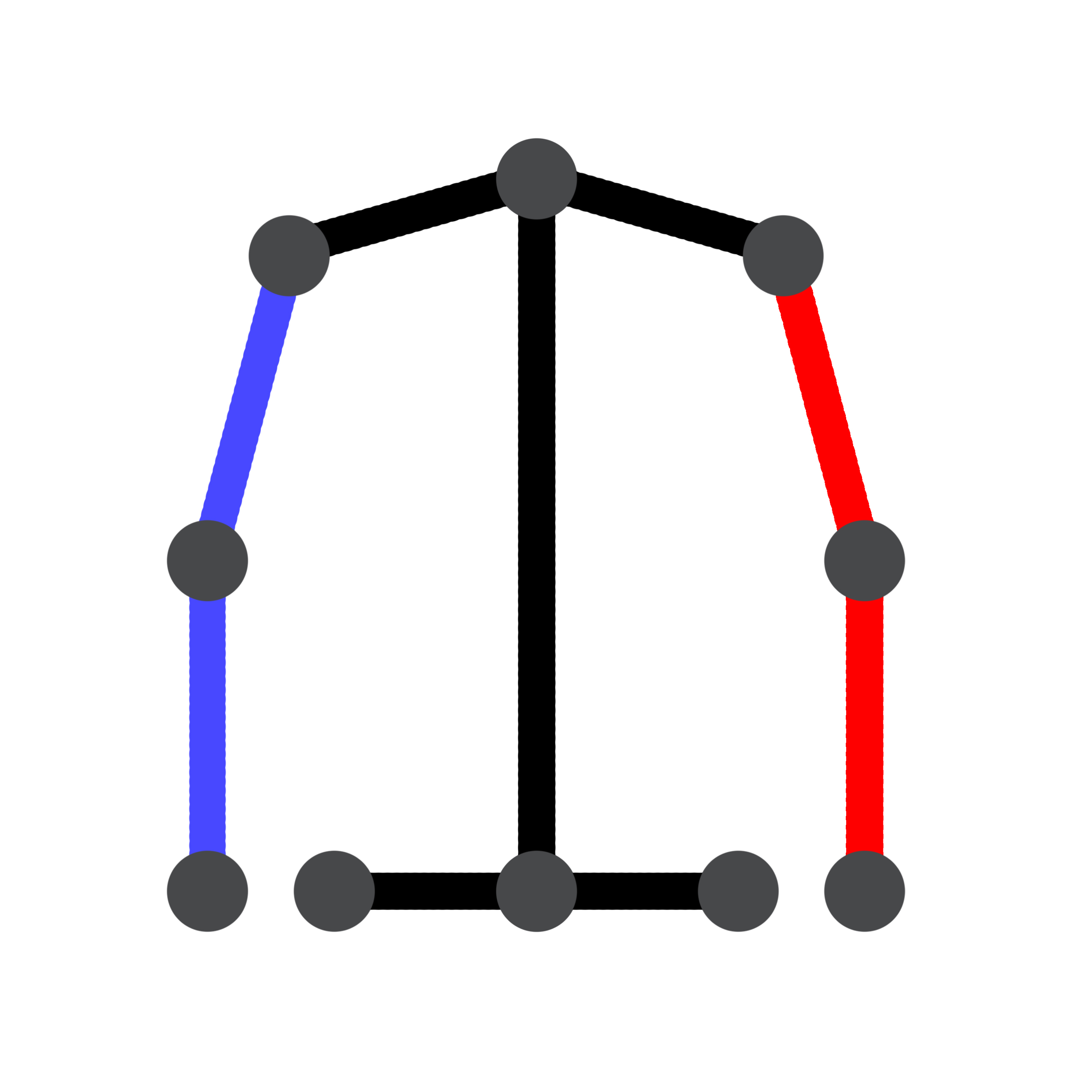}
        \caption{Front view}
        \label{fig:front}
    \end{subfigure}
    \caption{Poses that are indistinguishable without occlusion.}
    \label{fig:imp-occlusion}
\end{figure}

However, 2D stick figures omit important information contained in the image, most notably the depth information implied by occlusion \cite{i3d_pavlakos2018ordinal, 25d3d_zhou2019hemlets} \cref{fig:imp-occlusion}. In addition to losing occlusion information—which can be a problem even within one dataset—the successful application of ``lifting" depends on both z-score normalization \cite{i3d_sun2017compositional, 2d3d_martinez2017simple, 2d3d_wandt2019repnet, 2d3d_gcn_zhao2019semantic, 2d3d_gcn_zou2021modulated, 2d3d_hyp_li2019generating, 2d3d_hyp_wang2018drpose3d, 2d3d_hyp_sharma2019monocular} and some knowledge of camera viewing angle. Since both can differ significantly between datasets, 2D keypointing methods tend to perform poorly across datasets \cite{i3d_cd_wang2020predicting, 2d3d_gong2021poseaug}, likely because the elimination of occlusion introduces a bias that a network spuriously interprets as a non-existent prior. %(cf. \cref{tab: cd-mpjpe}). 

The viewpoint also plays a key role in understanding human pose, as distinguishing left from right is crucial to figuring out subject's orientation. 2D stick figures cannot preserve the relative ordering when the left and right keypoints overlap with each other \cref{fig:left-p}.

To improve cross-dataset performance, we need to: (1) avoid depending on any metric %(\eg the mean and standard deviation)
derived from the training set; (2) estimate the viewpoint accurately; and (3) avoid discarding occlusion information.

%
%\begin{enumerate}
%    \item avoid depending on any metric (\eg the mean and standard deviation) derived from the training set,
%    \item estimate the viewpoint accurately; and
%    \item avoid discarding occlusion information.
%\end{enumerate}
%

\begin{figure*}[t]
    \centering
    \includegraphics[width=1.0\linewidth]{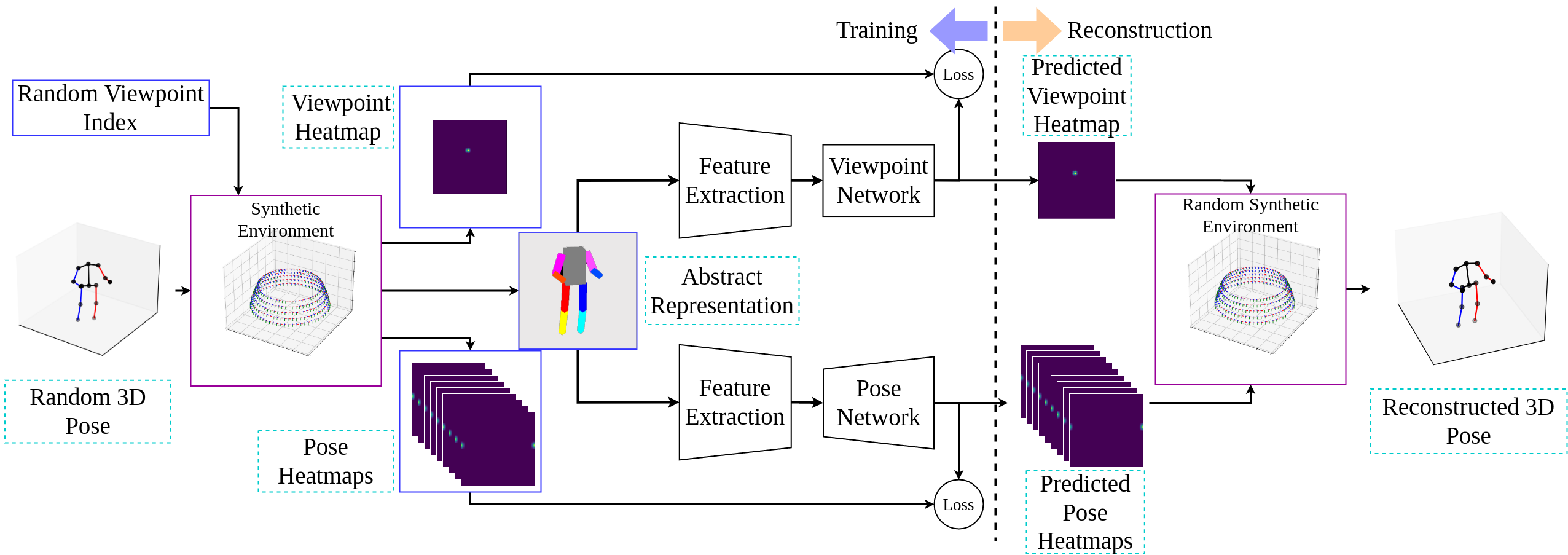}
    \caption{Overview of our approach. \textbf{Training}: First, we use a 3D pose and a random viewpoint to generate an abstract image, viewpoint heatmap, and pose heatmaps from the synthetic environment. The viewppoint and pose heatmaps are used as supervised training targets. Both the pose network, and the viewpoint network have a backbone feature extraction network. The feature extraction network takes in the synthetic image and feeds extracted features to the pose and viewpoint network. We optimize the L2 loss on the output of viewpoint and pose network with heatmaps generated from the synthetic environment. \textbf{Reconstruction}: The trained network takes synthetic image as input and generates a viewpoint and pose prediction heatmaps. These heatmaps are passed into a random synthetic environment to reconstruct a 3D pose.}
    \label{fig:overview}
\end{figure*}

Another hindrance to cross-dataset performance is this: the most common measure of error in ``pose'' estimation is not pose error at all, but {\em position} error—which is implicitly tied to z-score normalization. To successfully allow cross-dataset application, we must devise error measures that are truly about {\em pose}, rather than about {\em position}.

To simultaneously solve all of these problems, we propose to train on a huge multitude of synthetic images of opaque, solid-body “robot” humans across a huge dataset of real human poses, and taken from virtually all plausible camera viewpoints. We address the viewpoint bias by proposing a novel viewpoint encoding scheme that creates a 1-to-1 mapping between the camera viewpoint and the input image; a similar 1-to-1 encoding defines the pose. Both encodings support fully-convolutional training. Using the synthetic image as input to two networks, we train one for viewpoint and another for pose. At inference time, we take the predicted viewpoint and pose from the abstract image to reconstruct a new 3D pose. Since reconstruction does not ensure the correct forward facing direction of the subject, the ground-truth target pose will be related to the reconstructed pose by a rotation which can be easily accounted for to compare with other methods. \cref{fig:overview} shows a high level overview of our approach.

Our key observation is that the camera viewpoint as seen from the subject, and the subject's observed pose as seen from the camera, are {\em independent}: although they are intimately tied together in the sense that both are needed to fully reconstruct an abstract image, they answer completely separate questions. Namely: (1) the location of the camera as viewed from the subject is completely independent of the subject's pose; and (2) the pose of the subject is completely independent of where the camera is located. In the real world, these are simply two separate questions who's answers have absolutely no relation to each other---though if you want to reconstruct an abstract representation of the image as it was actually taken by the real camera, the answers to both are needed.

%For example, if a soldier is standing at attention and saluting with his right hand while his mother takes his photograph from directly behind him, his pose (``standing at attention while saluting'')---defined by his skeleton and unit vectors parallel with each limb---is an objective fact that is completely independent of the camera's location: his mother could move around while he remains motionless. Similarly, the camera's location (``directly behind him'') is an objective fact that is completely independent of his pose: he could have been standing ``at ease'', sitting in a chair, even lying prone on his side, but so long as he keeps his back to the camera, the camera's factual, objective location remains unchanged.

Note that humans can easily identify virtually any pose observed---so long as there is observable occlusion, which disambiguates many poses that would be indistinguishable without it. Thus, there exists a virtually 1-to-1 mapping between two-dimensional images, and three-dimensional poses. Similarly, the photographer can easily infer where she is with respect to the subject (``behind him'' or ``to his left'', etc.); thus, there is also a 1-to-1 mapping between the image and the subject-centered viewpoint.

Our method decomposes 3D human pose recognition into the above two orthogonal questions: (1) where is the camera in subject-centered coordinates, and (2) what is the observed pose (in terms of unit vectors along the subject's limbs) of the subject in camera coordinates as seen from the camera? Note that identical three-dimensional poses as viewed from different angles will change {\em both} answers, but combining the answers should always allow us to reconstruct a subject-centered pose that is the same in all cases.

This, then, is our ``secret sauce'': by incorporating occlusion information, we can independently train two fully convolutional systems: one that learns a 1-to-1 mapping between images and the subject-centered camera viewpoint, and another that learns a 1-to-1 mapping between images and camera-centered limb directions. The final ingredient is to train these two CNN's using a virtually unlimited set of ``abstract'' images of ``robots'' generated from randomly chosen camera viewpoints observing the ground-truth 3D joint locations of real humans in real poses, with occlusion. Given a sufficiently large (synthetic) dataset of abstract images, we are able to independently train two CNNs that reliably encode the two 1-to-1 mappings.

The key contributions of our paper are:
%\vspace{-1mm}
\begin{enumerate}
    \item We propose modeling the human body using solid, opaque, 3D shapes such as cylinders and rectangular blocks that preserve occlusion information and part-mapping;
    \vspace{-1mm} \item Novel viewpoint and pose encoding schemes, which facilitate learning a 1-to-1 mapping with input while preserving a spherical prior; and
    \vspace{-1mm} \item We improve state-of-the-art performance in cross-dataset benchmark without relying on dataset dependent normalization, and without sacrificing same-dataset performance.
\end{enumerate}

\section{Related Work}

\noindent{\textbf{Pose Estimation}} 3D pose estimation generally takes the form of a regression \cite{2d3d_chen20173d, 2d3d_martinez2017simple, 2d3d_gong2021poseaug, 2d3d_gcn_ci2019optimizing, 2d3d_gcn_zhao2019semantic, 2d3d_gcn_liu2020comprehensive, 2d3d_gcn_zeng2020srnet, 2d3d_gcn_zou2021modulated, 2d3d_hyp_li2019generating} with a fully-connected layer at the end, or a voxel-based approach \cite{25d3d_zhou2019hemlets, 2d3d_kin_mehta2017vnect, i3d_pavlakos2017coarse, i3d_pavlakos2018ordinal} with fully-convolutional supervision. The voxel-based method generally comes with a target space size of $w \times h \times d \times N$, where $w$ is the width, $h$ height, $d$ depth, and $N$ is the number of joints. On the other hand, the position regression requires some sort of training set dependent normalization (\eg z-score) \cite{i3d_sun2017compositional, 2d3d_martinez2017simple, 2d3d_wandt2019repnet, 2d3d_gcn_zhao2019semantic, 2d3d_gcn_zou2021modulated, 2d3d_hyp_li2019generating, 2d3d_hyp_wang2018drpose3d, 2d3d_hyp_sharma2019monocular}. Both the graph convolution based approach \cite{2d3d_gcn_ci2019optimizing, 2d3d_gcn_zeng2020srnet, 2d3d_gcn_zhao2019semantic, 2d3d_gcn_zou2021modulated, 2d3d_gcn_liu2020comprehensive} and hypothesis generation approach \cite{2d3d_hyp_li2019generating, 2d3d_hyp_wang2018drpose3d, 2d3d_hyp_sharma2019monocular} rely on z-score normalization to improve same-dataset, and most crucially, cross-dataset performance. To address missing depth information, Pavlakos \etal \cite{ i3d_pavlakos2018ordinal} propose including ordinal relations in training. To a similar cause, Zhou \etal \cite{25d3d_zhou2019hemlets}, proposed a heatmap triplet based intermediate representation per part. Our pose encoding scheme is fully-convolutional and has a smaller memory footprint in contrast to a voxel-based approach (by a factor of $d$) and does not depend on normalization parameters from training set.

\noindent{\textbf{Part Based Approach}} Kundu \etal \cite{part_kundu2020self} applies an unsupervised part-guided approach to 3D pose estimation. From an image, they generate part-segmentation with the help of intermediate 3D pose and a 2D part dictionary. Our approach, in contrast, is supervised and uses a part-mapped synthetic image to predict viewpoint and 3D pose.

\noindent{\textbf{Viewpoint}} Viewpoint estimation generally boils down to regressing some form of $(\theta, \phi)$ \cite{vp_ghezelghieh2016learning}, rotation matrix \cite{vp_Zimmermann_2017_ICCV, vp_Wandt_2022_CVPR}, or quaternions \cite{i3d_cd_wang2020predicting}. Regardless of the approach, everyone agrees on viewpoint estimation relative to the subject. However, relative subject rotation makes it harder to estimate viewpoint accurately.

\noindent{\textbf{Relation to previous work}}
We train on synthetically-generated images of ``robots” whose pose is derived from ground-truth 3D human poses. The “robot” has opaque, 3D limbs that are uniquely color-coded (implicitly defining a part-map). We call 2D projection of such a representation an ``abstract image”, in the sense that it contains the minimum information required to completely describe a human pose. (There already exist methods to convert real images into abstract ones similar to ours \cite{cdcl_lin2020cross}; incorporating this into our method is a work-in-progress.) Most existing approaches use regression on either 3D joint positions or voxels. Our own early tests showed that the former performs extremely badly across datasets when the same z-score parameters are used for both training and test sets, and improves only marginally if the normalization parameters are independently computed for both training and test sets (which is infeasible in the field, but is reported in \cref{tab: cd-mpjpe} below). Conversely, voxel regression presents a trade-off in performance vs. memory footprint as voxel resolution is increased. Our pose encoding (1) does not require training set dependent normalization, (2) takes much less memory than a voxel-based representation (by a factor of d), and (3) being heatmap-based, it integrates well in a fully-convolutional setup. Finally, most methods encode the viewpoint using a rotation matrix, sine and cosines, or quaternions; all of these methods suffer from a discontinuous mapping at $2\pi$. In contrast, our method avoids discontinuities by training a network on a Gaussian heat-map of viewpoint (or pose) that wraps around at the edge; as a result, the network learns that the heatmap can essentially be viewed as being on a cylinder.
\section{Method}
\subsection{Synthetic Environment}
\label{subsec: syn_env}
\begin{figure}[t]
    \centering
    \begin{subfigure}{0.48\linewidth}
        \centering
        \includegraphics[width=0.95\columnwidth]{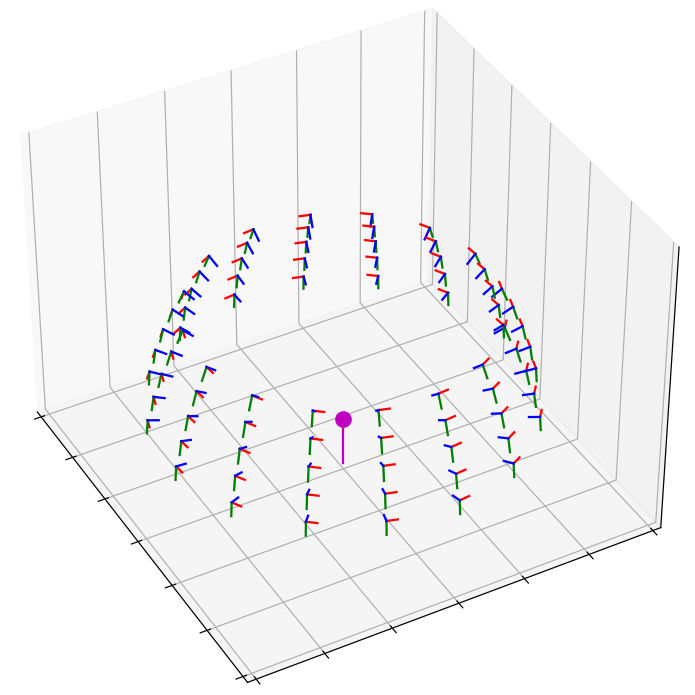}
        \caption{Synthetic Environment}
        \label{fig:syn_env_example}
    \end{subfigure}
    \begin{subfigure}{0.48\linewidth}
        \centering
        \includegraphics[width=0.95\columnwidth]{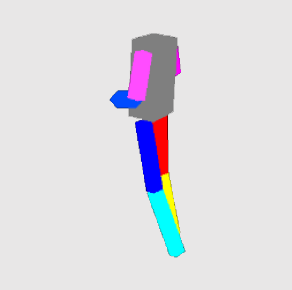}
        \caption{Abstract Image}
        \label{fig:abs_shape_1}
    \end{subfigure}
    \caption{(a) Setup of a synthetic environment. Cameras are arranged spherically and all point to $\vec{f}$ (magenta dot near the center). (b) If left forearm and femur is colored blue, it becomes easy to figure out where the subject is facing, whereas a “stick figure” representation that loses occlusion information has difficulty determining the “front facing” direction.}
\end{figure}

The synthetic environment can be thought of as a room full of cameras arranged systematically and all pointing to the same fixed point at the center of the room. We define $\vec{\mathrm{T}} \in \mathbb{R}^{\mathrm{X} \times \mathrm{Y} \times 3}$, the translation/position of the cameras in $\mathrm{X}$ columns and $\mathrm{Y}$ rows. The fixed point is defined as, $\vec{f} = \frac{c}{\mathrm{XY}} \sum \vec{\mathrm{T}}$, where $c < 0.5$. The scaling constant ($c$) helps the top cameras to point down from above, which is necessary during training to account for a wide variety of possible camera positions at test time.

As shown in \cref{fig:syn_env_example}, each camera is related to the room via a rotation matrix, $\mathrm{R} \in \mathbb{R}^{\mathrm{X} \times \mathrm{Y} \times 3 \times 3}$. We compute the look vector as $\vec{l}_{ij} = \vec{f} - \vec{\mathrm{T}}_{ij}$ for camera $(i,j)$ and take a cross-product with $-\hat{z}$ as the up vector to compute the right vector $\vec{r}$, all of which are fine-tuned to satisfy orthonormality by a series of cross-products. Refer to \cref{sec:implementation}, for predefined values.

\subsection{Abstract Shape Representation}
\label{subsec: abs_repr}

To ensure that occlusion information is clear in our synthetic images, our robot’s 8 limbs and torso use 9 easily-distinguishable, high-contrast colors (\cref{fig:abs_shape_1}). The 3D joint locations define the endpoints of the appropriate limbs (\eg, the upper and lower arm limbs meet at the 3D location of the elbow). In contrast to related work that used unsupervised training on rigid transformations of 2D spatial parts \cite{part_kundu2020self}, our method analytically generates a synthetic image with opaque limbs and torso intersecting at the appropriate 3D joint locations.

\begin{figure}[h]
    \centering
    \begin{subfigure}{0.48\linewidth}
        \centering
        \includegraphics[width=0.85\columnwidth]{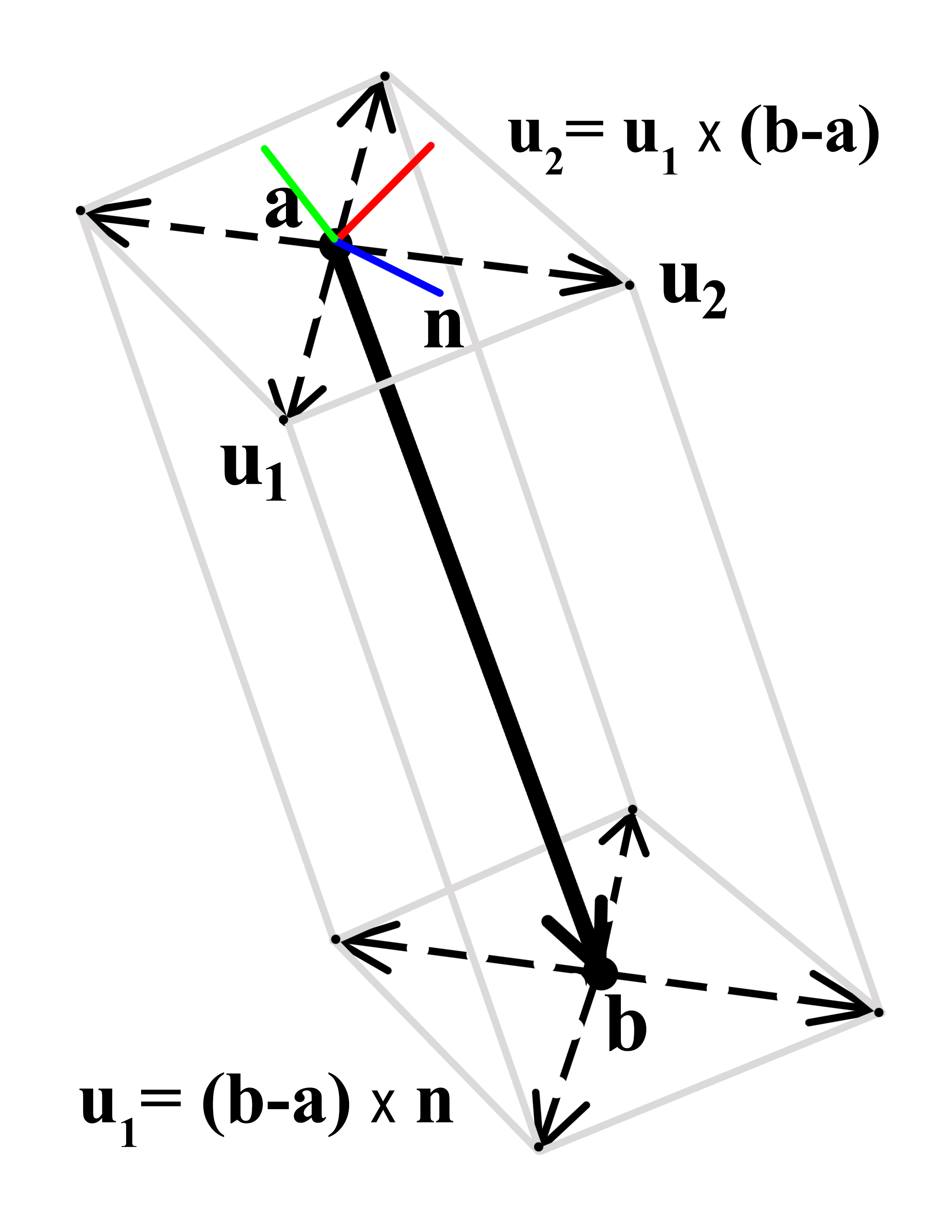}
        \caption{Limb Generation}
        \label{fig:cube_from_vec}
    \end{subfigure}
    \begin{subfigure}{0.48\linewidth}
        \centering
        \includegraphics[width=0.85\columnwidth]{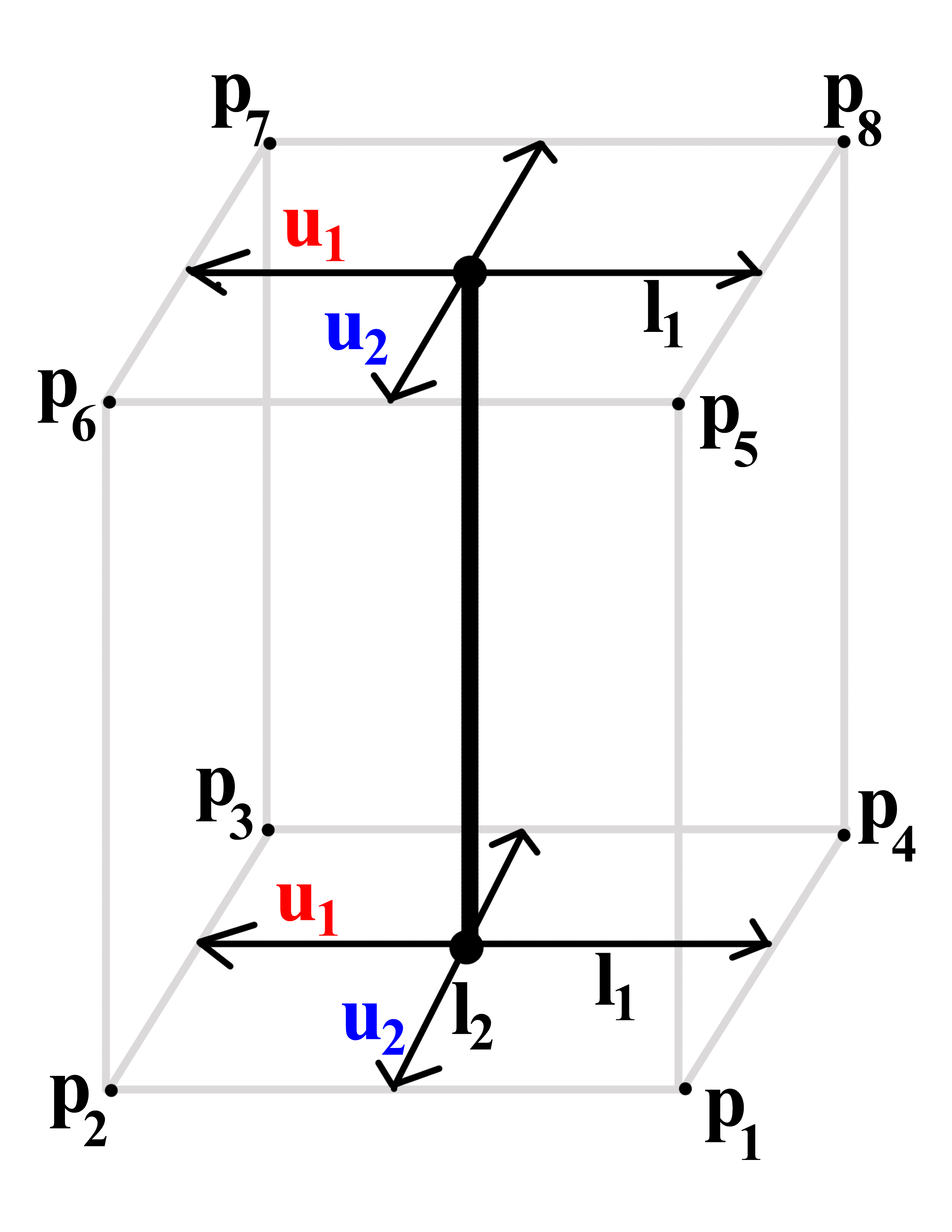}
        \caption{Torso Generation}
        \label{fig:cube_from_axes}
    \end{subfigure}
    \caption{(a) Limb generation from a vector; (b) Torso generation from right and forward vectors}
\end{figure}

\begin{figure*}[t]
    \begin{subfigure}{0.23\linewidth}
        \centering
        \includegraphics[width=0.95\columnwidth]{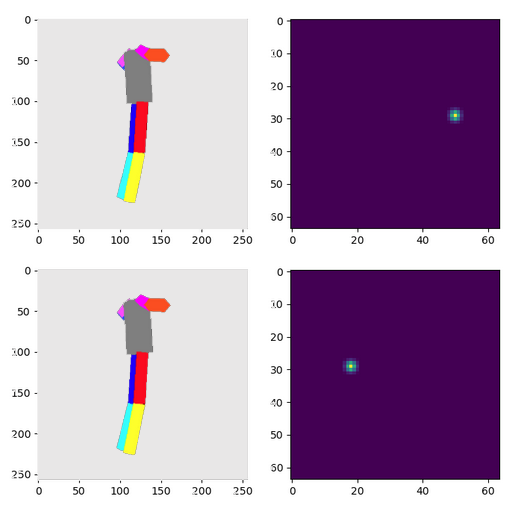}
        \caption{Na\"{i}ve approach}
        \label{fig:ve_naive}
    \end{subfigure}
    \begin{subfigure}{0.23\linewidth}
        \centering
        \includegraphics[width=0.95\columnwidth]{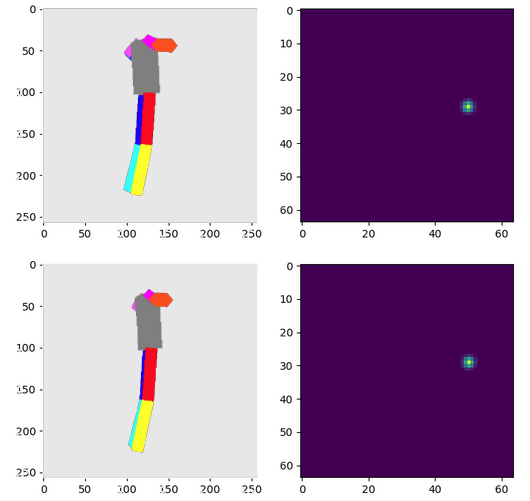}
        \caption{Rotation invariant approach}
        \label{fig:ve_improved}
    \end{subfigure}
    \begin{subfigure}{0.23\linewidth}
        \centering
        \includegraphics[width=0.95\columnwidth]{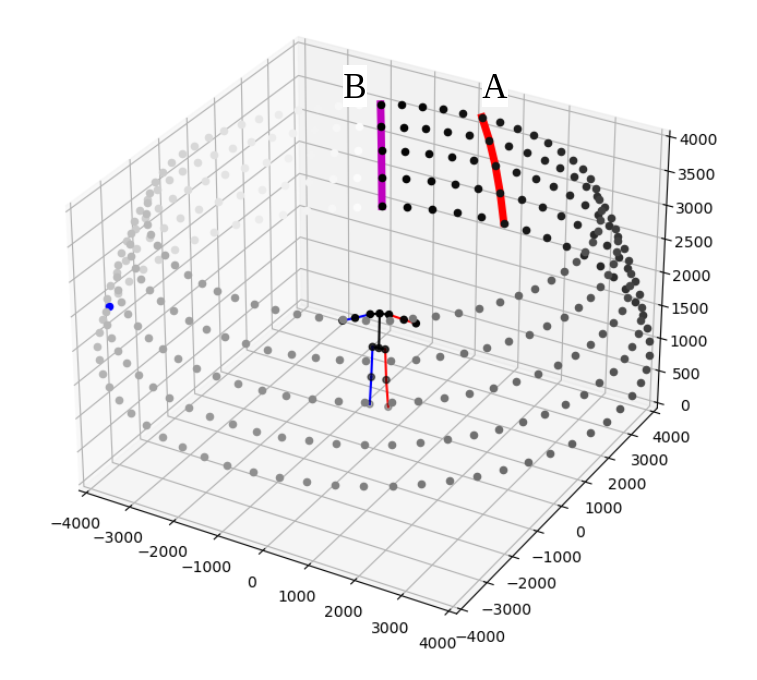}
        \caption{Seam Lines}
        \label{fig:ve_seams}
    \end{subfigure}
    \begin{subfigure}{0.23\linewidth}
        \centering
        \includegraphics[width=0.75\columnwidth]{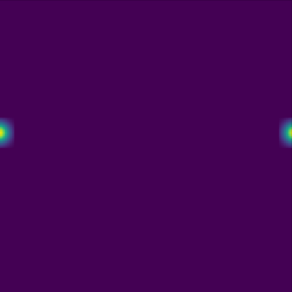}
        \caption{Wrapped Gaussian heatmap}
        \label{fig:gauss_hmap_wapred}
    \end{subfigure}
    \caption{(a) Na\"{i}ve approach of encoding viewpoint. As can be seen, for a rotated subject, we have same image but different viewpoint encoding. (b) Rotation invariant approach makes sure we have the same encoding if the image is same even if the subject is rotated. (c) Seam lines after computing cosine distances. Camera indices (black=0, white=63), rotate with subject. Seam line A (red) is the original starting point of the indices. Seam line B (purple) is the new starting point consistent with subject's rotation. (d) A Gaussian heatmap warped horizontally. }
    \label{fig:viewencoding}
\end{figure*}

Ours limbs and torso are formed by cuboids with orthogonal edges formed via appropriate cross-products; limbs (\cref{fig:cube_from_vec}) have a long axis (a to b) along the bone with a square cross-section, while the torso (\cref{fig:cube_from_axes}) is longest along the spine and has a rectangular cross-section. While the limb cuboid is generated from a single vector (a to b), the torso is generated with the help of body centered coordinate system \cite{i3d_cd_wang2020predicting}.

%For the limbs, we obtain an auxiliary vector, $\vec{u}_1 = (\vec{b}-\vec{a}) \times \vec{n}$, where $\vec{n} \in \{ \hat{x}, \hat{y}, \hat{z} \}$ and not parallel to the main vector. A series of corss-product with $\vec{u}_1$ and $\vec{b}-\vec{a}$, we get our control points for limb cuboid.

%For the torso (\cref{fig:cube_from_axes}), we start by extracting the body centered coordinate system \cite{i3d_cd_wang2020predicting}. We set the auxiliary vectors, $\vec{u}_1$ to the right vector and $\vec{u}_2$ to the forward vector. From here, we can use scalar multiplication and vector addition to find 4 end-points $(p_1, p_2, p_3, p_4)$ around hip $\vec{h}$ and 4 end-points $(p_5, p_6, p_7, p_8)$ around neck $\vec{n}$.

\begin{algorithm}
\caption{Abstract Shape Generation}\label{alg:shape_gen}
\KwData{$\mathrm{P}_{cam} \in \mathbb{R}^{3 \times N}$, $f_{cam}$, $c_{cam}$, $\text{colors} \in \mathbb{R}^{N \times 3}$}
\KwResult{$\mathcal{A}$}
$ \mathrm{X}_{3D} \gets \text{compute\_cuboids($\mathrm{P}_{cam}$)}$\;
$ \mathrm{X}_{2D} \gets \text{project\_points($\mathrm{X}_{3D}$, $f_{cam}$, $c_{cam}$)}$\;
$ \mathrm{H}_{2D} \gets \text{QHull($\mathrm{X}_{2D}$)}$\;
$ \mathrm{D} \gets \text{sort(compute\_distance($\mathrm{P}_{cam}$))}$\;
$ \mathcal{A} \in \mathbb{R}^{W \times H \times 3}$\;
\For{i \text{in descending order of } $\mathrm{D}$}{
    $\text{poly}_i$ $\gets$ extract\_polygon($\mathrm{H}_{2D}[i]$)\;
    $\mathcal{A}[\text{poly}_i] \gets \text{Colors}_i$ 
}
\end{algorithm}

Let all the endpoints be compiled in a matrix $\mathrm{X}_{3D} \in \mathbb{R}^{3 \times N}$, where $N$ is number of parts. We project this points to 2D $\mathrm{X}_{2D} \in \mathbb{R}^{2 \times N}$ using the focal length $f_{cam}$ and camera center $c_{cam}$ (predefined for a synthetic room). Using the QHull algorithm \cite{qhull}, we compute the convex hull of the projected 2D points for each limb. We compute the Euclidean distance between each part's midpoint and the camera. Next, we iterate over the parts in order of longest distance, extract the polygon from hull points, and assign limb colors.

\subsection{Viewpoint Encoding}
\label{subsec: vp_enc}
Our goal is to obtain an encoding that ensures a 1-to-1 mapping from the input image to relative camera position, and learns the spherical mapping of the room.

We show the problem of a na\"{i}ve approach with encoding azimuth ($\theta$) and elevation ($\phi$) of the camera relative to the subject as a Gaussian heatmap on a 2D matrix in \cref{fig:ve_naive}. Two different cameras can generate the same image resulting in two different viewpoint heatmap.

%The most na\"{i}ve approach at the first objective would be to get the azimuth ($\theta$) and elevation ($\phi$) relative to the subject and regress it directly. As a baseline approach we map the $\theta$ and $\phi$ value on a discretized matrix. We can start from any camera row/column in our synthetic environment and assign indices. If we pick camera $(i, j)$, on the matrix, at index $(i,j)$, we create a Gaussian heatmap. Using a heatmap allows us to perform a fully-convolutional training. However, this na\"{i}ve approach would not ensure a 1-to-1 mapping as shown in \cref{fig:ve_naive}: if the person rotates and a different camera generates the same image, we will have two viewpoint encodings for the same input image.

To address this, we take the idea of wrapping a matrix in a cylindrical formation. We call the edge where the matrix edges meet a \emph{seam line} (\cref{fig:ve_seams}). The key intuition behind our approach is to define an encoding, where the seam line is always at the back of the subject—ie., opposite to their forward vector. This ensures the coordinates on the matrix always stay in a fixed point related to the subject's orientation.

We compute the cosine distance between subject's forward vector $\vec{\mathrm{F}}_s$ projected onto xy-plane $\vec{\mathrm{F}}_{sp}$, and camera's forward vector $\vec{\mathrm{F}}_c$ and place the seam line (index 0 and 63 of the matrix) directly behind the subject. \cref{fig:ve_improved} reflects the improvement from \cref{fig:ve_naive}. Note for the same input, we have same viewpoint encoding now.

To learn a spherical mapping, we have to make the network understand the spherical positioning of the cameras. In general, a normal heatmap-based regression will clip the Gaussian at the border of the matrix. On the contrary, we allow the Gaussian heatmaps in the matrix to wrap around at the boundaries — corresponding to the seam line. Let
\begin{equation}
    \label{eqn: vp_gauss}
    \mathcal{G}(x, y, \mu_x, \mu_y) = \exp^{-\frac{(x-\mu_x)^2+(y-\mu_y)^2}{2\sigma^2}}
\end{equation}
be the formula for a Gaussian value at $(x, y)$ around $(\mu_x, \mu_y)$. Then the heatmap is:
\begin{equation}
    \mathcal{H}^{v}[i, j] = 
    \begin{cases}
        \mathcal{G}(j, i, \mu_x, \mu_y), &\text{if } |\mu_x - j| < W_k\\
        \mathcal{G}(j-I_w, i, \mu_x, \mu_y), &\text{if } |j-I_w-\mu_x| < W_k\\
        \mathcal{G}(j+I_w, i, \mu_x, \mu_y), &\text{if } |\mu_x-I_w-j| < W_k
    \end{cases}
    \label{eqn:vp_hmap}
\end{equation}
where $(\mu_x, \mu_y)$ is the index of the viewpoint in our rotated synthetic room. $I_w$ is the image size, and $W_k$ is the kernel width. \cref{alg:rotate_cam_arr} is used to rotate the camera indices in the synthetic room to ensure the camera position is consistent with the subject.

\begin{algorithm}
    \caption{Rotate Camera Array}
    \KwData{$\mathrm{S}_e \text{ (Synthetic Environment)}$, $\text{$\vec{\mathrm{F}}_s$ (Subject Forward Vector)}$}
    \KwResult{$\mathrm{T}^\prime$, $\mathrm{R}^\prime$}
    $\vec{\mathrm{F}}_c \gets \mathrm{S}_e\text{.camera\_forwards}$\;
    $\vec{\mathrm{F}}_{sp} \gets \vec{\mathrm{F}}_s - (\vec{\mathrm{F}}_s \cdot \hat{z})\hat{z}$\;
    $\mathrm{D} \gets \vec{\mathrm{F}}_c \cdot \vec{\mathrm{F}}_{sp}$\;
    $\mathcal{S} \gets \text{argmax } \mathrm{D}$\;
    $\mathcal{I} \gets \mathrm{S}_e\text{.original\_index\_array}$\;
    $\mathcal{I}_r \gets \text{rotate\_index\_array($\mathcal{I}$, $\mathcal{S}$)}$\;
    $\mathrm{T} \gets \mathrm{S}_e\text{.camera\_position}$\;
    $\mathrm{R} \gets \mathrm{S}_e\text{.camera\_rotation}$\;
    $\mathrm{T}^\prime \gets \mathrm{T}[\mathcal{I}_r]$\;
    $\mathrm{R}^\prime \gets \mathrm{R}[\mathcal{I}_r]$\;
    \label{alg:rotate_cam_arr}
\end{algorithm}

This encodes the camera position in subject space and addition of Gaussian heatmap relaxes the area for network to optimize on (i.e. picking an almost approximate neighboring camera).

\subsection{Pose Encoding}
\label{subsec: pose_enc}

\begin{figure*}[t]
    \centering
    \begin{subfigure}{0.24\linewidth}
        \centering
        \includegraphics[width=0.9\columnwidth]{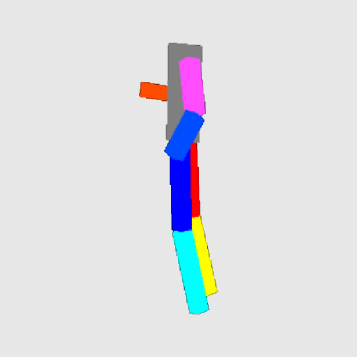}
        \caption{Input image}
        \label{fig:inp_img_2}
    \end{subfigure}
    \begin{subfigure}{0.24\linewidth}
        \centering
        \includegraphics[width=0.9\columnwidth]{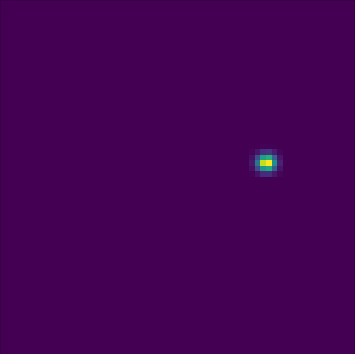}
        \caption{Predicted viewpoint}
        \label{fig:pred_vp_2}
    \end{subfigure}
    \begin{subfigure}{0.24\linewidth}
        \centering
        \includegraphics[width=0.9\columnwidth]{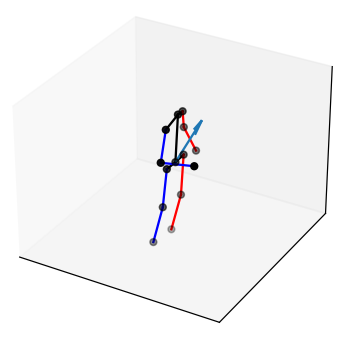}
        \caption{Predicted pose}
        \label{fig:pred_pose_2}
    \end{subfigure}
    \begin{subfigure}{0.24\linewidth}
        \centering
        \includegraphics[width=0.9\columnwidth]{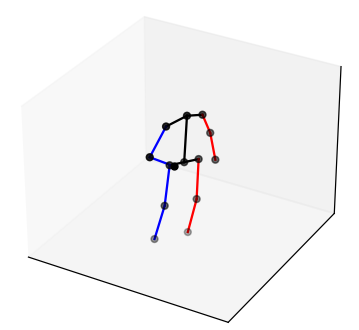}
        \caption{Ground truth pose}
        \label{fig:gt_pose_2}
    \end{subfigure}
    \caption{A sample output from our network. (a) is the synthetic image fed into the network. (b) the predicted viewpoint heatmap. (c) is the reconstructed pose from the pose and viewpoint heatmaps. The arrow shooting out the pose's left indicates the camera was left of the subject. (d) is the ground-truth 3D pose, which is  the reconstructed pose related with a rotation.}
    \label{fig: quiver_sample}
\end{figure*}

We decompose the pose into bone vectors $\mathcal{B}_r$, and bone lengths $\mathrm{B}_r$, both relative to parent joint. Let the synthetic environment's selected camera rotation matrix be $\mathrm{R}_{ij}$, and  $\mathcal{B}_{ij} = {\mathrm{R}_{ij}}^\prime\mathcal{B}_r$ be the bone vectors in $\mathrm{R}_{ij}$'s coordinate space. Then, we normalize the spherical angles ($\theta$, $\phi$) of $\mathcal{B}_{ij}$ from range $[-180, 180]$ to range $[0, 127]$. Note that this encoding is not dependent on any normalization of the training—and by implication, is also independent of any normalization of the test set. We now have $(\theta, \phi)$ normalized in a $128 \times 128$ grid. We take a similar approach to viewpoint encoding and allow the Gaussian heatmap generated around the matrix locations to wrap around the boundaries. Only difference is in viewpoint, we only needed to account for horizontal warping. Here, we account for both vertical and horizontal wraping. For joint $i$ and $k_1, k_2 \in [-\frac{W_k}{2}, \frac{W_k}{2}]$, 
\begin{equation}
    \mathcal{H}^p_i[h, g] = \mathcal{G}(k_1, k_2, 0, 0)
\end{equation}
where $h = \mu_y+k_2 (\text{mod $I_w$})$ and $g = \mu_y+k_1 (\text{mod $I_w$})$. Thus, we have another heatmap-based encoding for the pose. This encoding $\mathcal{H}^p \in \mathbb{R}^{128 \times 128 \times N}$, where N is the number of joints. \cref{fig:gauss_hmap_wapred} shows a wrapped version of the heatmap.

\subsection{Pose Reconstruction}

Since the camera viewpoint is encoded in a subject-based coordinate system, the first step of pose reconstruction is to transform the camera’s position from subject-centered coordinates to world coordinates. Let $\hat{\mathcal{H}}^v$ and $\hat{\mathcal{H}}^p$ be the output of viewpoint and pose network respectively. The non-maxima suppression on $\hat{\mathcal{H}}^v$ yields camera indices $(\hat{i},\hat{j})$, and spherical angles $(\hat{\theta}, \hat{\phi})$ from $\hat{\mathcal{H}}^p$. In an arbitrary synthetic room with an arbitrary seam line, we pick a subject forward vector, $\vec{\mathrm{F}}_s$ parallel to the seam line. Let the rotation matrix of camera at $(\hat{i},\hat{j})$ relative to $\vec{\mathrm{F}}_s$ be $\mathrm{R}_{\hat{i}\hat{j}}$. We obtain the Cartesian unit vectors $\mathcal{B}_{\hat{i}\hat{j}}$ from $(\hat{\theta}, \hat{\phi})$ and the relative pose in world space by, $\mathcal{B}_d = \mathrm{R}_{\hat{i}\hat{j}} \mathcal{B}_{\hat{i}\hat{j}}$. Then, we apply depth first traversal on $\mathcal{B}_d$ starting from the origin to reconstruct the pose using the bone lengths stored in our synthetic environment. 

%Note that this process reconstructs the \textbf{pose} in terms of the relative \emph{angles} of the limbs defined from an arbitrary co-ordinate system rather than the \emph{position} of the joints; and the pose is decoupled from the position of the \textbf{camera} in the input image. As hypothesis suggests thus far and also backed by experimental results later on, the predicted pose will be related to the actual pose by a rotation alignment only.

% this in turn makes a specification of joint ``positions’’ unnecessary since our pose is camera-independent—as it should be, because human poses exist even in the absence of a camera.

In \cref{fig: quiver_sample}, we show unseen test output from our actual network. Specifically, in \cref{fig:pred_pose_2}, note how the reconstructed pose is rotated from the ground-truth pose in \cref{fig:gt_pose_2}. The arrow shooting out from the subject's left in \cref{fig:pred_pose_2}, indicates the relative position of the camera when the picture was taken. 
%We marked the right leg with red and yellow color and the left with blue and teal. With this information, we can verify this is indeed the correct pose modulo a rotation in world-space.
\section{Implementation}
\label{sec:implementation}

For our experiment, we calculated the average bone lengths from the H36M dataset's training set \cite{h36m}. The viewpoint was discretized into $5 \times 64$ indices, and encoded into a $64 \times 64$ matrix. The $5$ rows span within $[25, 30]$ row range in the heatmap matrix. The fixed point scalar in our synthetic environment was set to $0.4$, the radius set to $5569$ mm. In principle, this setup could easily be extend to include cameras covering the entire sphere in order to, for example, account for images of astronauts floating in ISS as viewed from {\em any} angle. The pose was first normalized to fall in range $[0, 128]$ to occupy a $13 \times 128 \times 128$ matrix. Since we are using a $14$ joint setup, $13$ is the number of bones.

We trained two separate networks for pose and viewpoint. Both of these networks use HRNet \cite{hrnet_sun2019deep} as feature extraction module. The pose network consists two Convolution and Batch Normalization block pairs, followed by a transposed Convolution to match the output size of $128 \times 128$. All the convolution block use a $3 \times 3$ kernel with padding and stride set to 1. The final transposed convolution uses stride 2 and outputs a $13 \times 128 \times 128$ size tensor. For viewpoint estimation, we apply only one Convolution and Batch Normalization pair on the output of HRNet. The final stage is a regular convolution block that shrinks the output channel to 1 and outputs a $1 \times 64 \times 64$ size tensor.

We used a batch size of $64$ during training. Since our target is heatmap, we applied the standard L2 loss. We used Adam \cite{kingma2014adam} as our optimizer with learning rate set to $1\times10^{-3}$. The viewpoint network was run for $200$ epochs lasting $5$ days, and the pose network was run for $100$ epochs lasting $2.5$ days using RTX 3090. The pose network was stopped early since it hit an optimum.
\section{Experiments}
\subsection{Datasets and Evalution Metrics}
\iffalse
\begin{figure}
    \centering
    \begin{subfigure}{0.45\linewidth}
        \includegraphics[width=0.8\columnwidth]{figs/hip_shallow_3dpw.png}
        \caption{Shallow hip (3DPW)}
        \label{fig:shallow_hip}
    \end{subfigure}
    \begin{subfigure}{0.45\linewidth}
        \includegraphics[width=0.8\columnwidth]{figs/hip_correct_3dpw.png}
        \caption{Corrected hip}
        \label{fig:corrected_hip}
    \end{subfigure}
    \caption{Shallow hip problem in 3DPW and SURREAL. (a) Shows the shallow hip and also the hip bone is above the two femurs. We fix it to be consistent with GPA and H36M dataset. (b) After we correct hip.}
    \label{fig: shallow-hip-issue}
\end{figure}
\fi

\noindent \textbf{Human3.6M Dataset (H36M)} \cite{h36m} contains $15$ actions performed $7$ actors in a $4$ camera setup. In our experiment, we \emph{only} take the 3D pose in world co-ordinate space to train our network. We follow the standard protocol to keep subject 1, 5, 6, 7, 8 for training, and 9, 11 for testing.

\noindent \textbf{Geometric Pose Affordance Dataset (GPA)}\cite{gpa} has 13 actors interacting with a rich 3D environment and performing numerous actions. Only used for cross-dataset testing.

\setlength{\tabcolsep}{2pt}
\begin{table*}[t]\centering
    \scriptsize
    \begin{tabular}{lrrrrrrrrrrrrrrrrr}\toprule
        \textbf{Protocol \#1} &\textbf{Direct.} &\textbf{Discuss} &\textbf{Eating} &\textbf{Greet} &\textbf{Phone} &\textbf{Photo} &\textbf{Pose} &\textbf{Purch.} &\textbf{Sitting} &\textbf{SittingD.} &\textbf{Smoke} &\textbf{Wait} &\textbf{WalkD.} &\textbf{Walk} &\textbf{WalkT.} &\textbf{Avg.} \\\midrule
        Moreno \etal \cite{2d3d_moreno20173d} (CVPR'17) &69.54 &80.15 &78.2 &87.01 &100.75 &102.71 &76.01 &69.65 &104.71 &113.91 &89.68 &98.49 &79.18 &82.4 &77.17 &87.3 \\
        Chen \etal \cite{2d3d_chen20173d} (CVPR'17) &71.63 &66.6 &74.74 &79.09 &70.05 &93.26 &67.56 &89.3 &90.74 &195.62 &83.46 &71.15 &85.56 &55.74 &62.51 &82.72 \\
        Martinez \etal \cite{2d3d_martinez2017simple} (ICCV'17) &51.8 &56.2 &58.1 &59 &69.5 &78.4 &55.2 &58.1 &74 &94.6 &62.3 &59.1 &65.1 &49.5 &52.4 &62.9 \\
        Yang \etal \cite{2d3d_yang20183d} (CVPR'18) &51.5 &58.9 &50.4 &57 &62.1 &65.4 &49.8 &52.7 &69.2 &85.2 &57.4 &58.4 &43.6 &60.1 &47.7 &58.6 \\
        Sharma \etal \cite{2d3d_hyp_sharma2019monocular} (ICCV'19) &48.6 &54.5 &54.2 &55.7 &62.6 &72 &50.5 &54.3 &70 &78.3 &58.1 &55.4 &61.4 &45.2 &49.7 &58 \\
        Zhao \etal \cite{2d3d_gcn_zhao2019semantic} (CVPR'19) &47.3 &60.7 &51.4 &60.5 &61.1 &49.9 &47.3 &68.1 &86.2 &55 &67.8 &61 &42.1 &60.6 &45.3 &57.6 \\
        Pavlakos \etal \cite{i3d_pavlakos2018ordinal} (CVPR'18) &48.5 &54.4 &54.4 &52 &59.4 &65.3 &49.9 &52.9 &65.8 &71.1 &56.6 &52.9 &60.9 &44.7 &47.8 &56.2 \\
        Ci \etal \cite{2d3d_gcn_ci2019optimizing} (ICCV'19) &46.8 &52.3 &44.7 &50.4 &52.9 &68.9 &49.6 &46.4 &60.2 &78.9 &51.2 &50 &54.8 &40.4 &43.3 &52.7 \\
        Li \etal \cite{2d3d_hyp_li2019generating} (CVPR'19) &43.8 &48.6 &49.1 &49.8 &57.6 &61.5 &45.9 &48.3 &62 &73.4 &54.8 &50.6 &56 &43.4 &45.5 &52.7 \\
        Martinez \etal \cite{2d3d_martinez2017simple} (GT) (ICCV'17) &37.7 &44.4 &40.3 &42.1 &48.2 &54.9 &44.4 &42.1 &54.6 &58 &45.1 &46.4 &47.6 &36.4 &40.4 &45.5 \\
        Zhao \etal \cite{2d3d_gcn_zhao2019semantic} (GT) (CVPR'19) &37.8 &49.4 &37.6 &40.9 &45.1 &41.4 &40.1 &48.3 &50.1 &\textbf{42.2} &53.5 &44.3 &40.5 &47.3 &39 &43.8 \\
        Zhou \etal \cite{i3d_zhou2016deep} (ICCV'19) &34.4 &42.4 &36.6 &42.1 &38.2 &\textbf{39.8} &34.7 &40.2 &45.6 &60.8 &39 &42.6 &42 &\textbf{29.8} &31.7 &39.9 \\
        Gong \etal \cite{2d3d_gong2021poseaug} (GT) (CVPR'21) &- &- &- &- &- &- &- &- &- &- &- &- &- &- &- &38.2 \\\midrule
        Ours &\textbf{30.39} &\textbf{34.28} &\textbf{31.03} &\textbf{32.63} &\textbf{33.24} &46.74 &\textbf{33.15} &\textbf{33.60} &\textbf{42.72} &57.20 &\textbf{35.60} &\textbf{38.15} &\textbf{34.49} &32.04 &\textbf{31.38} &\textbf{36.44} \\
        \bottomrule
    \end{tabular}
\caption{Quantitative comparisions of MPJPE (Protocol \#1) between the ground truth 3D pose and reconstructed 3D pose after a rotation. The best score in each column is marked bold. Our method outperforms in every action except Sitting Down and Walk. This shows the strength of our approach. Lower is better.}\label{tab: prot-1}
\end{table*}

\begin{table*}[h]\centering
\scriptsize
    \begin{tabular}{lrrrrrrrrrrrrrrrrr}\toprule
        \textbf{Protocol \#2} &\textbf{Direct} &\textbf{Discuss} &\textbf{Eating} &\textbf{Greet} &\textbf{Phone} &\textbf{Photo} &\textbf{Pose} &\textbf{Purch} &\textbf{Sitting} &\textbf{SittingD} &\textbf{Smoke} &\textbf{Wait} &\textbf{WalkD.} &\textbf{Walk} &\textbf{WalkT.} &\textbf{Avg} \\\midrule
        Moreno \etal (CVPR'17) &66.1 &61.7 &84.5 &73.7 &65.2 &67.2 &60.9 &67.3 &103.5 &74.6 &92.6 &69.6 &71.5 &78 &73.2 &74 \\
        Martinez \etal (ICCV'17) &39.5 &43.2 &46.4 &47 &51 &56 &41.4 &40.6 &56.5 &69.4 &49.2 &45 &49.5 &38 &43.1 &47.7 \\
        Li \etal (CVPR'19) &35.5 &39.8 &41.3 &42.3 &46 &48.9 &36.9 &37.3 &51 &60.6 &44.9 &40.2 &44.1 &33.1 &36.9 &42.6 \\
        Ci \etal (ICCV'19) &36.9 &41.6 &38 &41 &41.9 &51.1 &38.2 &37.6 &49.1 &62.1 &43.1 &39.9 &43.5 &32.2 &37 &42.2 \\
        Pavlakos \etal (CVPR'18) &34.7 &39.8 &41.8 &38.6 &42.5 &47.5 &38 &36.6 &50.7 &56.8 &42.6 &39.6 &43.9 &32.1 &36.5 &41.8 \\
        Sharma \etal (ICCV'19) &35.3 &35.9 &45.8 &42 &40.9 &52.6 &36.9 &35.8 &43.5 &51.9 &44.3 &38.8 &45.5 &29.4 &34.3 &40.9 \\
        Zhou \etal (ICCV'19) &\textbf{21.6} &\textbf{27} &29.7 &28.3 &\textbf{27.3} &\textbf{32.1} &\textbf{23.5} &30.3 &\textbf{30} &\textbf{37.7} &\textbf{30.1} &\textbf{25.3} &34.2 &\textbf{19.2} &\textbf{23.2} &\textbf{27.9} \\\midrule
        Ours &24.74 &29.09 &\textbf{27.36} &\textbf{27.69} &28.69 &40.47 &28.26 &\textbf{29.74} &38.05 &54.14 &31.42 &31.76 &\textbf{30.89} &26.45 &25.91 &31.64 \\
        \bottomrule
    \end{tabular}
    \caption{Quantitative comparison of PA-MPJPE (Protocol \#2) between the ground truth 3D pose and reconstructed 3D pose. The best score in each column is marked bold. Our method falls short of the state-of-the-art by ~3 millimeters. Lower is better.}\label{tab: prot-2}
\end{table*}

\noindent \textbf{3D Poses in the Wild Dataset}\cite{3dpw} is an ``in-the-wild" dataset with complicated poses and camera angles. Only used for cross-dataset testing.

\noindent \textbf{SURREAL Dataset} \cite{varol17_surreal} is one of the largest synthetic datasets with renderings of photorealistic humans. Only used for cross-dataset testing.

\noindent \textbf{Evaluation Metrics} We report Mean Per Joint Position Error (MPJPE) in millimeters, we call this Protocol \#1 and MPJPE after Procrustes Alignment (PA-MPJPE) as Protocol \#2, following convention. Since the reconstructed pose is related with the ground truth by rotation only, we report it under Protocol \#1. Further, PA-MPJPE reduces the error since the reconstruction uses preset bone-lengths. This becomes prominent in cross-dataset benchmarks.

\begin{figure*}[th]
    \centering
    \begin{subfigure}{0.16\linewidth}
        \centering
        \includegraphics[width=0.55\linewidth]{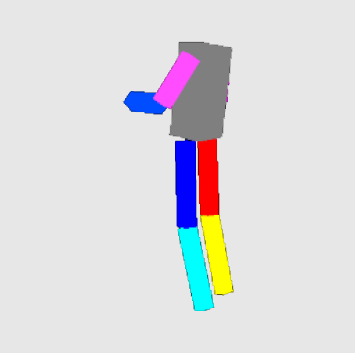}
        \caption{Input Image}
    \end{subfigure}
    \begin{subfigure}{0.16\linewidth}
        \centering
        \includegraphics[width=0.7\linewidth]{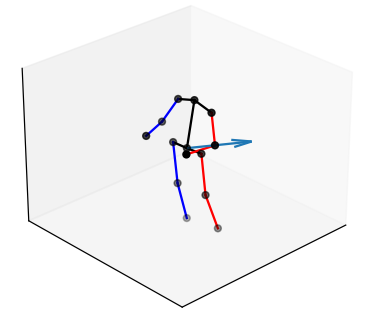}
        \caption{Prediction}
    \end{subfigure}
    \begin{subfigure}{0.16\linewidth}
        \centering
        \includegraphics[width=0.75\linewidth]{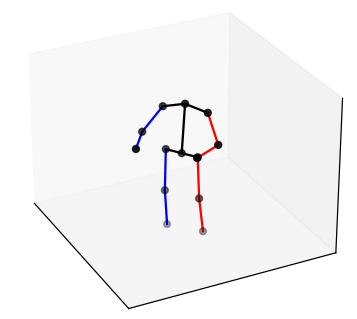}
        \caption{Ground Truth}
    \end{subfigure}
    \begin{subfigure}{0.16\linewidth}
        \centering
        \includegraphics[width=0.55\linewidth]{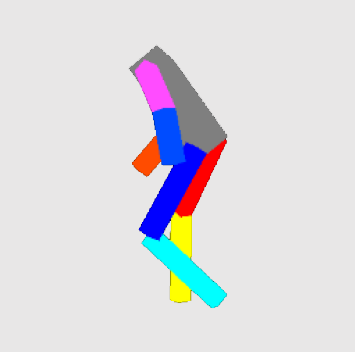}
        \caption{Input Image}
    \end{subfigure}
    \begin{subfigure}{0.16\linewidth}
        \centering
        \includegraphics[width=0.75\linewidth]{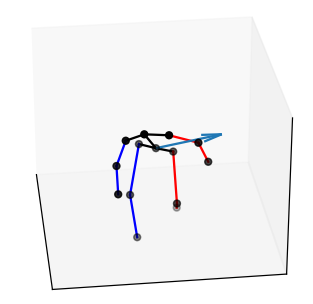}
        \caption{Prediction}
    \end{subfigure}
    \begin{subfigure}{0.16\linewidth}
        \centering
        \includegraphics[width=0.75\linewidth]{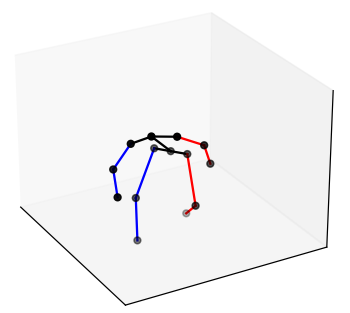}
        \caption{Ground Truth}
    \end{subfigure}
    \begin{subfigure}{0.16\linewidth}
        \centering
        \includegraphics[width=0.55\linewidth]{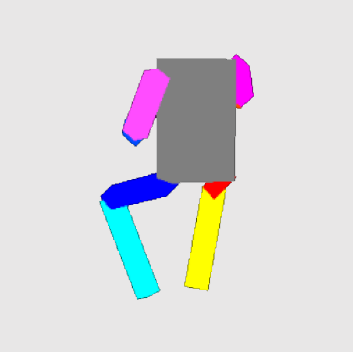}
        \caption{Input Image}
    \end{subfigure}
    \begin{subfigure}{0.16\linewidth}
        \centering
        \includegraphics[width=0.75\linewidth]{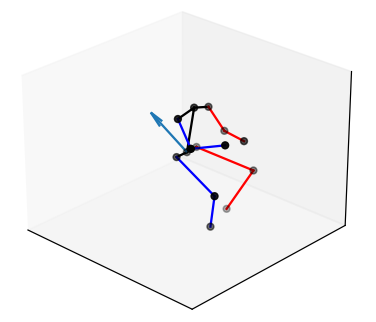}
        \caption{Prediction}
    \end{subfigure}
    \begin{subfigure}{0.16\linewidth}
        \centering
        \includegraphics[width=0.75\linewidth]{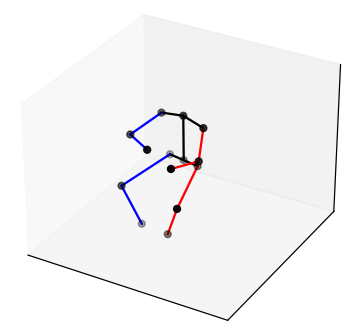}
        \caption{Ground Truth}
    \end{subfigure}
    \begin{subfigure}{0.16\linewidth}
        \centering
        \includegraphics[width=0.55\linewidth]{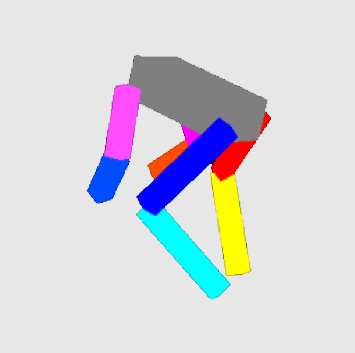}
        \caption{Input Image}
    \end{subfigure}
    \begin{subfigure}{0.16\linewidth}
        \centering
        \includegraphics[width=0.75\linewidth]{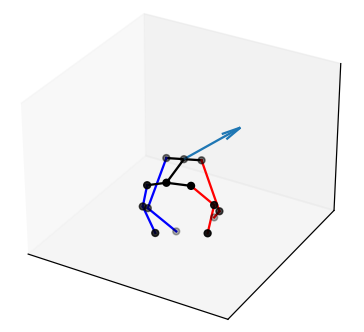}
        \caption{Prediction}
    \end{subfigure}
    \begin{subfigure}{0.16\linewidth}
        \centering
        \includegraphics[width=0.75\linewidth]{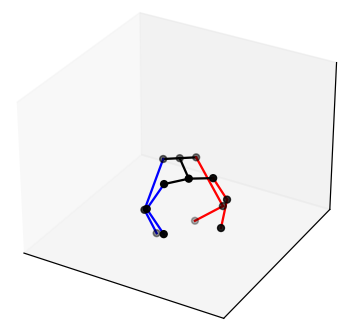}
        \caption{Ground Truth}
    \end{subfigure}
    \caption{Qualitative results on H36M dataset. Note how the blue arrow indicator always shows the relative camera position where the image was taken from.}
    \label{fig: qual}
\end{figure*}

\subsection{Evaluation on H36M Dataset}

We trained on 3D poses taken from H36M dataset. At training time, on each iteration, the poses are paired with a random sample of viewpoints from synthetic environemt to generate synthetic images. We do not use camera from H36M dataset during training. In testing, we use the camera configuration provided with the dataset to generate test images. We report the results in \cref{tab: prot-1} and \cref{tab: prot-2}. We outperform the state-of-the-art in all actions except ``Sitting Down" and ``Walk". Specifically, ``Sitting Down" is still a challenging task for our viewpoint encoding scheme because it relies on the projection of the forward vector. Leveraging a joint representation of spine and forward (which are orthogonal to each other), has potential to improve this encoding. We will address this in our future work. During reconstruction, we always use a preset bone-length. PA-MPJPE score on \cref{tab: prot-2}, which includes rigid transformation, accounts for bone-length variation and reduces the error even more.

\subsection{Cross-Dataset Generalization}

Cross-dataset results are hard to come by. To our best knowledge Wang \etal \cite{i3d_cd_wang2020predicting}, is the only work with an extensive cross-dataset analysis on $4$ datasets. To add more contender, we perform cross-dataset analysis on Martinez \etal \cite{2d3d_martinez2017simple} and Zhao \etal \cite{2d3d_gcn_zhao2019semantic}. We chose these two based on availability and adaptability of their code. Both of these methods rely on z-score normalization. The result presented for these two datasets are z-score normalized with testing set mean and standard deviation. This gives them an unfair advantage. Even after that, we still take the lead in cross-dataset performance. We show the results in MPJPE in \cref{tab: cd-mpjpe}.

Gong \etal \cite{2d3d_gong2021poseaug} reported cross-dataset performance on 3DPW dataset in PA-MPJPE. We have also included their result for comparison in \cref{tab: cd-pampjpe}. Again, we outperform the results by a significant margin. PA-MPJPE score again accounts for bone length discrepancy among datasets and reports much lower error in GPA, 3DPW, and SURREAL datasets compared to their MPJPE counterpart.

For this configuration, we have a network trained only with H36M poses. To test generalization capabilities, we render the images from GPA, 3DPW, and SURREAL dataset. For all the datasets, we made sure the subjects up vector in general is aligned with the z-direction of the world co-ordinate system. 3DPW and SURREAL's marker system introduces a shallow hip %(\cref{fig: shallow-hip-issue}) 
problem for all subjects, which we corrected with vector algebra.

\begin{table}[!htp]\centering
\scriptsize
\begin{tabular}{lrrrrr}\toprule
\textbf{Method} &\textbf{H36M} &\textbf{GPA} &\textbf{3DPW} &\textbf{SURREAL} \\\midrule
Martinez \etal \cite{2d3d_martinez2017simple}* &55.52 &117.37 &135.53 &108.63 \\
Zhao \etal \cite{2d3d_gcn_zhao2019semantic}* &53.59 &115.01 &154.3 &103.75 \\
Wang \etal \cite{i3d_cd_wang2020predicting} &52 &98.3 &124.2 &114 \\\midrule
Ours &\textbf{36.44} &\textbf{98.04} &\textbf{105.3} &\textbf{76.55} \\
\bottomrule
\end{tabular}
\caption{Cross-Dataset results on GPA, 3DPW, SURREAL in \textbf{MPJPE}. We take the lead across the board. Asterisk marks our own experiment. Note: the networks were trained on H36M}\label{tab: cd-mpjpe}
\end{table}

\begin{table}[!htp]\centering
\scriptsize
\begin{tabular}{lrrr}\toprule
\textbf{Method} & \textbf{GPA} &\textbf{3DPW} & \textbf{SURREAL} \\\midrule
Zhao \etal \cite{2d3d_gcn_zhao2019semantic} & - & 152.3 & -\\
Martinez \etal \cite{2d3d_martinez2017simple} & - & 145.2 & - \\
ST-GCN \cite{st-gcn} (1-Frame) & - & 154.3 & - \\
VPose \cite{Pavllo_2019_CVPR} (1-Frame) & - & 146.3 & -\\\midrule
Zhao \etal\cite{2d3d_gcn_zhao2019semantic} + Gong \etal\cite{2d3d_gong2021poseaug} & - & 140 & - \\
Martinez \etal \cite{2d3d_martinez2017simple} + Gong \etal\cite{2d3d_gong2021poseaug} & - & 130.3 & - \\
ST-GCN \cite{st-gcn} (1-Frame) + Gong \etal \cite{2d3d_gong2021poseaug} & - & 129.7 & - \\
VPose \cite{Pavllo_2019_CVPR} (1-Frame) + Gong \etal \cite{2d3d_gong2021poseaug} & - & 129.7 & - \\\midrule
Ours & \textbf{74.83} & \textbf{70.74} & \textbf{59.31}\\
\bottomrule
\end{tabular}
\caption{Cross-Dataset results on GPA, 3DPW, and SURREAL in \textbf{PA-MPJPE}. We show a performance improvement by almost a factor of two in all scenarios. Results from this table are taken from \cite{2d3d_gong2021poseaug}. Note: the networks were trained on H36M}\label{tab: cd-pampjpe}
\end{table}

\subsection{Qualitative Results}

\cref{fig: qual} shows the qualitative performance of our network on H36M. We see impressive viewpoint estimation indicated by the blue arrow on the second column of each test sample. This, indeed shows the accuracy and efficacy of our method on distangling viewpoint from pose.

\subsection{Ablation Study}

First, we show the pose encoding is independent from viewpoint in \cref{tab:pose_vp_ind}. For this, we set up three configurations using: (1) ground truth viewpoint and pose, (2) ground truth viewpoint and predicted pose, and (3) predicted viewpoint and pose. As shown in \cref{tab:pose_vp_ind}, we have a baseline error of 18.289 mm, which comes from the bone-length disparity. The second configuration increases the error by about 17 mm, since we are including predicted pose. The third configuration (including viewpoint prediction), increases the error by just 0.02mm -- 20 microns -- a completely negligible amount.. This indicates that the pose error is independent of the viewpoint - which is our goal.

\begin{table}
    \scriptsize
    \centering
    \begin{tabular}{lccc}\toprule
         \textbf{Experiment} & Configuration 1 & Configuration 2 & Configuration 3 \\
         \midrule
         \textbf{MPJPE (mm)}  & 18.289 & 35.489 & 35.507 \\\bottomrule
    \end{tabular}
    \caption{Results in MPJPE millimeters for configuration 1 (ground truth viewpoint and pose), configuration 2 (ground truth viewpoint and predicted pose), and configuration 3 (predicted viewpoint and pose)}
    \label{tab:pose_vp_ind}
\end{table}

\begin{figure}
    \centering
    \includegraphics[width=0.65\linewidth]{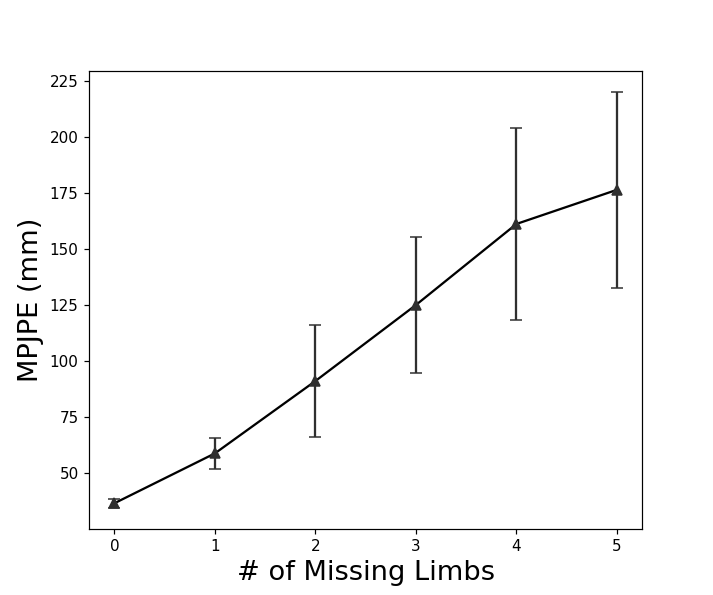}
    \caption{Error (MPJPE) vs number of missing parts. As more parts get missing from the input image, the error and uncertainty increases.}
    \label{fig:err_bar_abl}
\end{figure}

Next, we explore the impact on performance when the input image is missing some body parts. This worst-case scenario can arise if the part segmentation network completely misses one or many parts. To simulate this scenario, at the time of synthetic image generation, we randomly skip rendering a subset of the limbs. As expected from \cref{fig:err_bar_abl}, the error and uncertainty in prediction increases with the number of missing parts. This only shows the network cannot make an educated guess of what the pose would be if a limb suddenly disappears, since our network depends on occlusion to predict the pose in the first place.
\section{Conclusion}
We have shown that splitting pose lifting into the two parts “viewpoint prediction” and “pose prediction”, using a model trained on synthetic abstract images that include occlusion, gives far better cross-dataset results than using 2D keypoints, at an acceptable cost to same-dataset results. With a novel encoding scheme, we have shown how this method is on par with the state-of-the-art in same-dataset benchmark, while significantly advancing the bar on cross-dataset performance. We are working on a framework that will enable a seamless integration of our representation in any image-based model.

%%%%%%%%% REFERENCES
{\small
\bibliographystyle{ieee_fullname}
\bibliography{egbib}
}

\end{document}